\documentclass[conference]{IEEEtran}

\usepackage{times}   
\usepackage{mathptmx} 

\usepackage[ruled,vlined]{algorithm2e}
\IEEEoverridecommandlockouts
\usepackage{cite}
\usepackage{multirow}

\usepackage{amsmath,amssymb,amsfonts}
\usepackage{graphicx}
\usepackage{textcomp}
\usepackage{xcolor}
\def\BibTeX{{\rm B\kern-.05em{\sc i\kern-.025em b}\kern-.08em
    T\kern-.1667em\lower.7ex\hbox{E}\kern-.125emX}}
\begin{document}



\title{MOTIF-RF: \underline{M}ulti-template \underline{O}n-chip \underline{T}ransformer Synthesis \underline{I}ncorporating \underline{F}requency-domain Self-transfer Learning for \underline{RF}IC Design Automation
}

\author{
    Houbo He\textsuperscript{1*}, 
    Yizhou Xu\textsuperscript{1*}, 
    Lei Xia\textsuperscript{1}, 
    Yaolong Hu\textsuperscript{1},
    Fan Cai\textsuperscript{2}, 
    Taiyun Chi\textsuperscript{1} \\
    \textsuperscript{1}ECE Department, Rice University, Houston, TX, USA \quad
    \textsuperscript{2}Keysight Technologies, Santa Rosa, CA, USA \\
    \textsuperscript{*}Equally Credited Authors (ECAs)  \\
    \{hh68, yx103, lx27, yh72, taiyun.chi\}@rice.edu, fan.cai@keysight.com
}

\maketitle

\begin{abstract}
This paper presents a systematic study on developing multi-template machine learning (ML) surrogate models and applying them to the inverse design of transformers (XFMRs) in radio-frequency integrated circuits (RFICs). Our study starts with benchmarking four widely used ML architectures, including MLP-, CNN-, UNet-, and GT-based models, using the same datasets across different XFMR topologies. To improve modeling accuracy beyond these baselines, we then propose a new frequency-domain self-transfer learning technique that exploits correlations between adjacent frequency bands, leading to $\sim$$30\%\text{--}50\%$ accuracy improvement in the S-parameters prediction. 
Building on these models, we further develop an inverse design framework based on the covariance matrix adaptation evolutionary strategy (CMA-ES) algorithm. This framework is validated using multiple impedance-matching tasks, all demonstrating fast convergence and trustworthy performance. These results advance the goal of AI-assisted “specs-to-GDS” automation for RFICs and provide RFIC designers with actionable tools for integrating AI into their workflows.
\end{abstract}
%
%

\section{Introduction}

The rapid evolution of AI has sparked growing interest in AI-assisted design automation within the RFIC design community \cite{aspdac,KaushikNC,kaushik,PulseRF,Inp,hua2020,IMSTR,PACOSYT,LNA2025}. 
This emerging paradigm promises to significantly enhance design productivity compared with the conventional manual design flow and lower barriers to entry for fresh graduates and less-experienced designers. 

A critical distinction between RFIC and low-frequency analog IC design lies in the \emph{extensive use of passive networks} \cite{MWSCAS}.
These passive networks perform key circuit functions at RF, such as impedance matching, filtering, power combining or dividing, harmonic shaping, and more. Automating the design of passive networks is therefore a critical step towards realizing a fully automated RFIC design flow. 

\begin{figure}[t]
\centering
\includegraphics[trim=0.9 0.9 0.9 0.9, clip, width=3.5in]{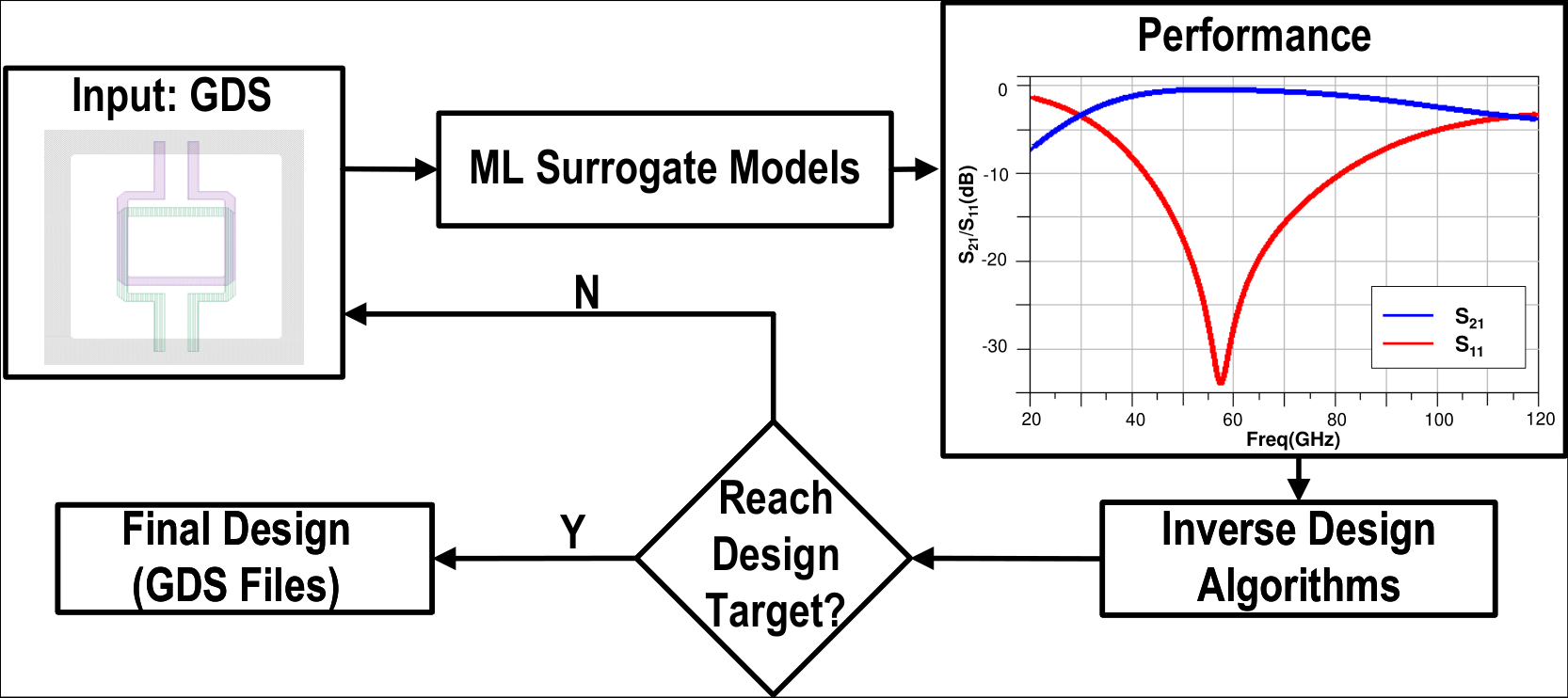}
\caption{“Specs to GDS” for AI-assisted design automation of XFMRs in RFICs, incorporating forward surrogate models and inverse design algorithms.}
\label{fig:framework}
\end{figure}

As shown in Fig. \ref{fig:framework}, the process of AI-assisted passive network design automation typically involves two stages. First, ML surrogate models are trained to replace time-consuming EM solvers used in the conventional design flow, enabling rapid yet accurate performance evaluation. Second, inverse design algorithms are applied, generating layouts that meet target specifications. This automated design process is also referred to as “specs to GDS” \cite{aspdac}.


For RFICs, on-chip transformers (XFMRs), i.e., magnetically coupled coils, are the most widely used components for passive network synthesis \cite{JohnLong,5GFEM,8shape}. While several recent studies have demonstrated the feasibility of building ML surrogate models to enable inverse design of XFMRs \cite{hua2020,IMSTR,PACOSYT}, they still face three major limitations: (1) Advanced ML architectures, such as Graph Transformers (GT), and the modeling of spectral continuity, remain largely underexplored in this context. (2) A comprehensive comparison of different ML models evaluated on \emph{the same XFMR dataset} is missing. As a result, it is difficult for RFIC designers to evaluate the trade-offs among different models and select the most suitable one for their design needs. (3) The prediction accuracy of existing models is still limited. This results in noticeable discrepancies between ML-predicted and EM-simulated circuit performance especially at high frequency, limiting the practical \emph{reliability} of applying such models for inverse design tasks.

This paper aims to address these gaps, and our three key contributions are summarized as follows. 

\begin{itemize}

    \item We offer a comprehensive, “handbook-style” evaluation of different ML surrogate models for XFMRs using the same dataset. Our goal is to provide RFIC designers, \emph{regardless of their level of AI expertise}, with a clear understanding of these models, and enable informed decisions when integrating AI into their workflows.

    \item We propose a new frequency-domain self-transfer learning technique that leverages spectral continuity when training surrogate models, achieving a significant improvement in model prediction accuracy of $\sim$$30\%\text{--}50\%$.


    \item Using these surrogate models, we further demonstrate reliable inverse design through multiple impedance matching examples, where both the ML-predicted results and the EM-simulated results of the generated layouts closely align with the target specifications.

\end{itemize}

The remainder of this paper is organized as follows. Section II presents the background of XFMRs in RFIC design and outlines our data collection process. Section III introduces four ML surrogate models investigated in this study, including MLP, CNN, UNet, and GT, along with the proposed self-transfer learning technique. Section IV presents a comprehensive model evaluation and comparison. Section V demonstrates the inverse design flow using multiple impedance-matching examples. Section VI concludes this paper.



\section{Preliminaries}



\begin{figure}[t]
\centering
\includegraphics[trim=40 10 70 70, clip, width=3.2 in]{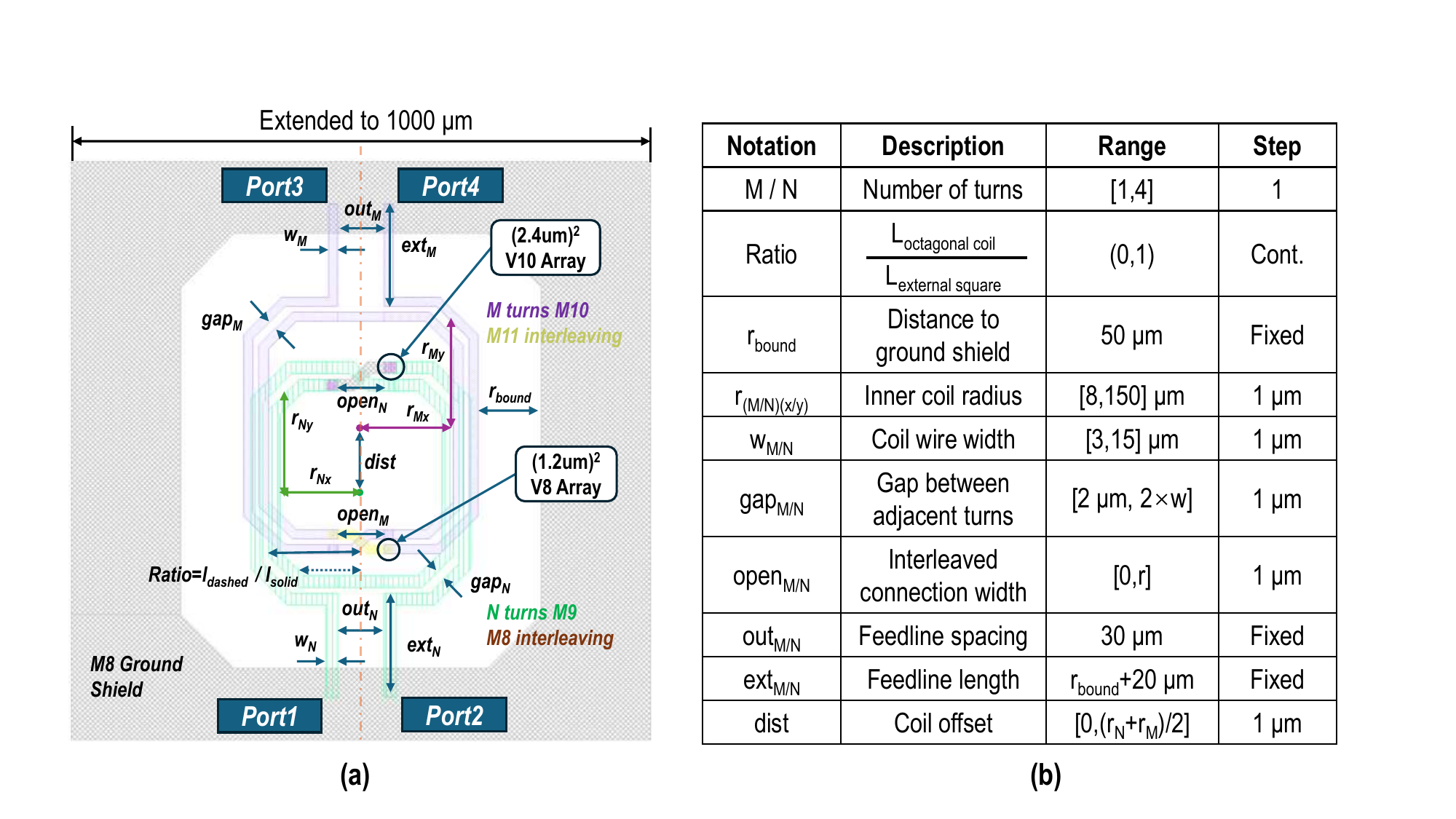}
\caption{(a) EM model of a generic $M$:$N$ XFMR and (b) its geometric parameters.}
\label{fig:nmgen}
\end{figure}

\begin{figure}[t]
\centering
\includegraphics[trim=100 40 100 40, clip, width=3.2 in]{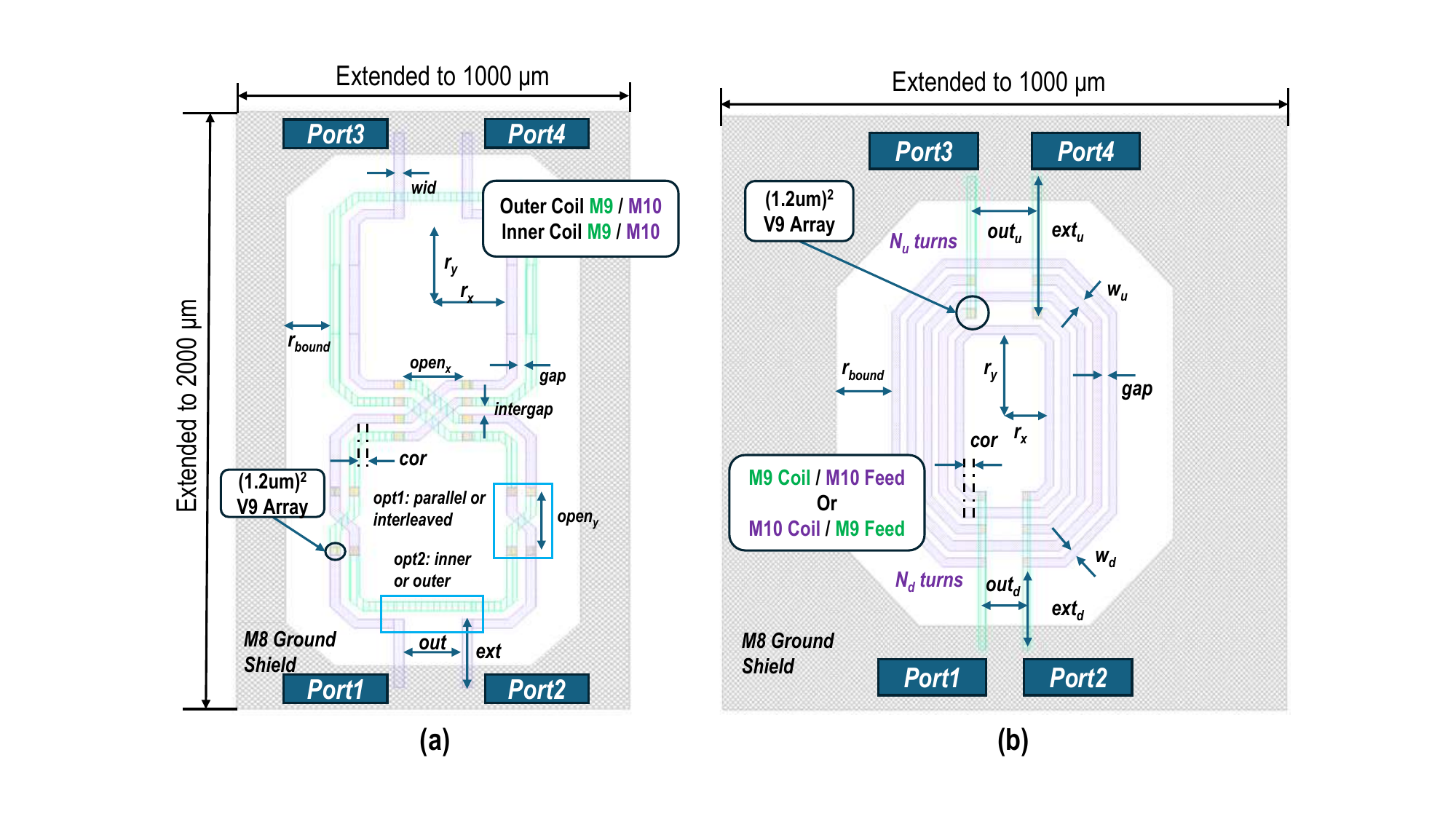}
\caption{(a) EM model of a generic 8-shaped-inductor-based XFMR. (b) EM model of a generic parallel-inductor-based XFMR. 
}
\label{fig:parallelgen}
\end{figure}

\subsection{XFMR Fundamentals and S-parameters}

A typical XFMR consists of two windings --  a primary winding and a secondary winding that are magnetically coupled. The EM model of a generic $M$:$N$ XFMR is shown in Fig. \ref{fig:nmgen}(a), with its geometric design parameters summarized in Fig. \ref{fig:nmgen}(b). 
Other than the generic $M$:$N$ XFMRs, other XFMR topologies have also been adopted by the RFIC industry. For example, the 8-shaped-inductor-based XFMR in Fig. \ref{fig:parallelgen}(a) offers significantly enhanced EMI immunity due to its rejection of adjacent magnetic fields \cite{8shape}, and the parallel-inductor-based XFMR in Fig. \ref{fig:parallelgen}(b) offers a stronger magnetic coupling with a higher passive efficiency at lower frequencies \cite{paraind}.

In the conventional manual design flow \cite{Yaolong_DAC}, XFMRs are analyzed using EM simulation software packages such as Cadence EMX, Keysight Momentum, or Ansys HFSS. These tools generate S-parameters, which are then imported into circuit simulators to evaluate circuit-level performance, such as inductance ($L$), quality factor ($Q$), impedance ($Z$), and loss over frequency \cite{XiaohanJSSC2023}. As EM simulation is computationally intensive, it has become a major productivity bottleneck in RFIC design. Although prior work has explored ML surrogate models to predict $L$ and $Q$ \cite{hua2020,IMSTR,PACOSYT}, the capability to directly predict broadband S-parameters remains underdeveloped. As \emph{S-parameters are the most fundamental representation of XFMRs}, from which nearly all performance characteristics can be derived \cite{pozar}, our work focuses on building surrogate models to directly predict S-parameters. Such capability is a critical prerequisite towards enabling efficient and reliable inverse design of the whole RFIC.

\subsection{Dataset Generation}

We prepared four datasets in this study: 5k samples for $1$:$1$ XFMRs; 16k samples for generic $M$:$N$ XFMRs, with 1k samples for each configuration where $M$ and $N$ range up to 4; 5k samples for parallel-inductor-based XFMRs; and 5k samples for 8-shaped-inductor-based XFMRs. The EM-simulated four-port S-parameters, spanning from 0.5/1 to 100/200 GHz with a 0.5/1-GHz step, are used as labels of the datasets (only the $1$:$1$ XFMR dataset includes data up to 200 GHz). This data collection process was done using Cadence EMX, taking about 80 hours in total with 16 threads.
Due to the layout symmetry and reciprocity of passive networks \cite{pozar}, only six out of the 16 S-parameters (i.e., $S_{11}$, $S_{12}$, $S_{13}$, $S_{14}$, $S_{33}$, $S_{34}$) are required to fully represent the four-port network. Each S-parameter is separated into real and imaginary components, resulting in a total dimensionality of 6$\times$2$\times$200 = 2,400.

It is important to note that all our datasets are generated based on the GlobalFoundries 22FDX process, a 22-nm CMOS SOI process that occupies significant market shares in automotive radar, IoT, 5G cellular communication, and other RF applications. This provides a realistic and industry-relevant basis for our study. This workflow could also be migrated to other technology nodes.



\section{Surrogate Modeling}
This section first outlines our ML surrogate models based on four different NN architectures: MLP, CNN, UNet, and GT. We then present the proposed frequency-domain self-transfer learning technique to enhance the model accuracy.

\subsection{ML Surrogate Models}
\subsubsection{Tabular-based Method Using Multi-layer Perceptron (MLP)}
Our first modeling approach leverages an MLP neural network, where the input features are tabular-based geometric parameters [as listed in Fig. \ref{fig:nmgen}(b)], and the outputs are the predicted real and imaginary components of the S-parameters. This method is straightforward and efficient, and serves as a baseline for evaluating more complex models.



\subsubsection{Image-based Methods Using CNN and UNet}

\begin{figure}[t]
\centering
\includegraphics[trim=1 1 1 1, clip, width=3.2 in, height = 2.6 in]{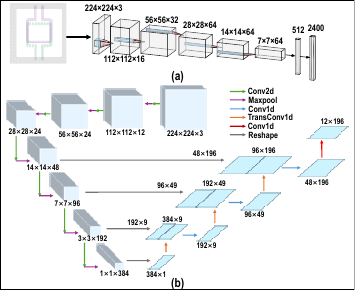}
\caption{(a) CNN-based and (b) UNet-based model architectures.}
\label{fig:CNN and Unet}
\end{figure}

The second approach treats XFMR layouts as images, enabling the application of image-based deep learning techniques such as CNN \cite{kaushik} and U-Net \cite{PulseRF}.
These models leverage the spatial structure inherent in the layouts to learn intricate geometric features that govern the EM behavior of XFMRs. As such, they are more advantageous in modeling complex multi-turn topologies and spatial dependencies that may be difficult to capture through tabular representations alone.
The architectures of these models are shown in Fig. \ref{fig:CNN and Unet}.

\subsubsection{Graph Transformer (GT)-based Method}
\begin{figure}[t]
\centering
\includegraphics[width=3 in]{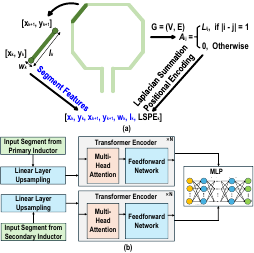}
\caption{(a) XFMR segment embedding. (b) GT-based model architecture.}
\label{fig:graph_transformer}
\end{figure}

The third approach leverages the recently developed GT architecture,
which excels in capturing topological properties of graph-structured data, making them particularly well-suited for circuit modeling.
We adopted an approach similar to \cite{SamSung}, which employs Laplacian Summation Positional Encoding (LSPE) to capture graph topological features. In our implementation, each metal segment of the XFMR is represented as a token, characterized by six geometric parameters and an LSPE encoding. These tokens form the input of the GT model, as shown in Fig. \ref{fig:graph_transformer}(a). Connections between adjacent nodes are represented by weighted edges, with edge weights determined by the lengths of the corresponding metal traces. This encoding scheme allows the model to distinguish between different XFMR sizes and shapes based on positional features. 
As shown in Fig. \ref{fig:graph_transformer}(b),
The tokens are vectorized, upsampled, and passed to the transformer encoder. This encoder follows the transformer framework \cite{Attention}, where each layer comprises a multi-head attention mechanism and a feedforward neural network. The encoded tokens for all XFMR segments are then concatenated to form a unified representation of the XFMR, followed by an MLP to map the extracted features to the corresponding S-parameters. 



\subsection{Proposed Frequency-domain Self-transfer Learning}

Conventional ML surrogate models for predicting S-parameters typically treat each frequency point independently, which ignores the \emph{spectral continuity governed by the underlying physics}. Although prior work such as STCNN \cite{STCNN} attempts to address this by adding a spectral convolutional layer, this layer is embedded as a hidden layer that lacks interpretability and no guarantee of capturing true frequency-domain correlations \cite{bengio2013representation}\cite{lipton2018mythos}.


In this work, we propose an iterative self-transfer learning technique, which explicitly leverages the correlation between adjacent frequency bands to improve model accuracy. Unlike conventional transfer learning or self-supervised learning methods \cite{self_learn}, \emph{our approach transfers knowledge learned from neighboring frequency bands within the same model}, allowing it to capture the spectral continuity more effectively. The detailed process is shown in Algorithm \ref{alg:frequency_transfer_learn}.
\begin{algorithm}[ht]
\small 
\caption{Iterative Self-transfer Learning Across Frequency Sub-bands}
\label{alg:frequency_transfer_learn}

\KwData{Number of sub-bands $N_{\text{band}}$, maximum frequency $f_{\mathrm{max}}$,
        number of transfer iterations $T$}

\setcounter{AlgoLine}{0}
Divide the full frequency range into $N_{\text{band}}$ sub-bands\;
Define each \texttt{sub-model[$i$]} to cover frequency range
$\left( \frac{(i-1) \cdot f_{\mathrm{max}}}{N_{\text{band}}},
        \frac{i \cdot f_{\mathrm{max}}}{N_{\text{band}}} \right]$\;
Train \texttt{sub-model[1]} using data from the first sub-band\;
Initialize transfer iteration counter $t = 1$\;

\While{$t \leq T$}{
  \textbf{Forward Transfer Training:}\;
  \For{$i = 1$ \textbf{to} $N_{\text{band}}-1$}{
    Initialize \texttt{sub-model[$i+1$]} with \texttt{sub-model[$i$]}\;
    Train \texttt{sub-model[$i+1$]} on sub-band $i+1$ data\;
  }

  \textbf{Backward Transfer Training:}\;
  \For{$i = N_{\text{band}}-1$ \textbf{down to} $1$}{
    Initialize \texttt{sub-model[$i$]} with \texttt{sub-model[$i+1$]}\;
    Train \texttt{sub-model[$i$]} on sub-band $i$ data\;
  }

  Increment transfer iteration counter $t \leftarrow t + 1$\;
}

\textbf{return} Final trained sub-models \texttt{sub-model[$1 \dots N_{\text{band}}$]}\;

\end{algorithm}





\section{Model Evaluation}
This section presents a comprehensive evaluation of all models introduced in Section III. All models were trained using the same loss function \cite{STCNN}, defined as
\begin{equation}\label{L_freq}
L_{freq}=\frac{1}{N}\sum_{n=1}^{N}\sqrt{\frac{1}{K}\sum_{k=1}^{K}(S_{n,k}-\hat{S}_{n,k})^{2}}
\end{equation}
where $k$ = 1,...,200 represents the frequency dimension, and $n$ = 1,...,12 represents the dimension of S-parameters.

A critical metric for comparing different models is accuracy measured by mean absolute error (MAE). To benchmark accuracy, we calculated the average MAE across all 12 S-parameter components at a single frequency point, denoted as MAE\textsubscript{freq}. Additionally, we report two forms of MAE\textsubscript{avg} averaged over a frequency range: MAE\textsubscript{avg,100G} (or MAE\textsubscript{avg,200G} for 1:1 XFMRs) and MAE\textsubscript{avg,2SRF}. The former is calculated over the full range of 100 (or 200) GHz, while the latter is calculated only up to 2$\times$ the self-resonant frequency (SRF) of each XFMR. As XFMRs are typically used below their SRF in practice \cite{haoisscc2025}, MAE\textsubscript{avg,2SRF} is a more relevant measure of modeling accuracy for real-world applications. The extended frequency range from SRF to 2$\times$SRF captures harmonic behaviors that can be useful for nonlinear circuits.

\subsection{Evaluation of Different Modeling Approaches}

We started with the comparison for $1$:$1$ XFMRs, with results summarized in Table \ref{evaluation_11}. These results lead to the following two key observations. (1) MLP- and GT-based models, which utilize geometric input features, outperform their image-based counterparts. This can be attributed to the relatively simple layout structure and low spatial complexity of 1:1 XFMRs. (2) Among models with similar model parameter counts ($\sim$2M), GT-based model achieves the lowest MAE\textsubscript{avg,2SRF} of 0.0103. This is likely attributed to GT's capability to extract latent structural features beyond basic coordinate inputs through graph positional encoding and attention mechanisms.

\begin{table}[t]
\caption{Evaluation on the $1$:$1$ XFMR Dataset}
\centering
\label{evaluation_11}
\begin{tabular}{|c|c|c|c|c|}
\hline
\parbox{20mm}{\centering \bfseries {Trained Model}}&\parbox{10mm}{\centering \bfseries MLP}&\parbox{10mm}{\centering \bfseries CNN}&\parbox{10mm}{\centering \bfseries UNet}&\parbox{10mm}{\centering \bfseries GT}\\
\hline
Params & 2.02M & 2.93M & 2.03M& 1.95M\\
\hline
$R^{2}$ & 0.978& 0.972 &0.978 & 0.987\\
\hline
MAE\textsubscript{avg,200G}& 0.0214 & 0.0252 & 0.0222&0.0164\\
\hline
MAE\textsubscript{avg,2SRF} & 0.0135 & 0.0151 & 0.0144&0.0103\\
\hline

\end{tabular}
\end{table}


A similar comparison was made for generic $M$:$N$ XFMRs, with results summarized in Table \ref{evaluation_mn}. Compared to $1$:$1$ XFMRs, the increased design space and spatial complexity of $M$:$N$ XFMRs allow image-based methods, especially the UNet-based model, to catch up with and even outperform the baseline MLP-based model, particularly when the parameter counts are increased to reach saturated MAE. This demonstrates the advantage of image-based methods in modeling complex layouts with larger degrees of freedom.

\begin{table}[t]
\caption{Evaluation on the Generic $M$:$N$ XFMR Dataset}
\centering
\label{evaluation_mn}
\begin{tabular}{|c|c|c|c|c|c|}
\hline
\parbox{20mm}{\centering \bfseries {Trained Model}}&\parbox{8mm}{\centering \bfseries MLP}&\parbox{8mm}{\centering \bfseries CNN}&\parbox{8mm}{\centering \bfseries UNet}&\parbox{8mm}{\centering \bfseries UNet}&\parbox{8mm}{\centering \bfseries GT}\\
\hline
Params & 2.02M & 2.93M & 2.03M&3.61M& 3.53M\\
\hline
$R^{2}$ & 0.886& 0.835 &0.887&0.911 &0.890 \\
\hline
MAE\textsubscript{avg,100G} & 0.0638 & 0.0803 & 0.0619&0.0513&0.0614\\
\hline
MAE\textsubscript{avg,2SRF} & 0.0244 & 0.0333 & 0.0252&0.0180&0.0252\\
\hline
\end{tabular}
\end{table}

Results for parallel-inductor-based and 8-shape-inductor-based XFMRs are presented in Table \ref{evaluation_parallel} and Table \ref{evaluation_8shape}, respectively. For these two pre-defined templates, MLP-based approach achieves the lowest MAE\textsubscript{avg}. 
The relatively complicated contours of XFMR metal segments makes the GT-based modeling less effective.

Additionally, we observe a limitation regarding the implemented linear transformer architecture. While GT-based model performs well when operating on relatively small geometrically encoded inputs, it struggles with more complex graphs, especially when extended to include additional layout encodings beyond single-layer metal traces such as \emph{multi-layer routings and vias}. This observation is consistent with prior literature \cite{Rampasek2022} \cite{DwivediBresson2021}, which note that transformer models demand large datasets and carefully engineered structural encodings \cite{Ying2021} \cite{Dwivedi2022} to scale well beyond simple or medium-sized graph scenarios.

\begin{table}[t]
\vspace{-1.5em}
\caption{Evaluation on the Parallel-inductor-based XFMR Dataset}
\centering
\label{evaluation_parallel}
\begin{tabular}{|c|c|c|c|c|}
\hline
\parbox{20mm}{\centering \bfseries {Trained Model}}&\parbox{10mm}{\centering \bfseries MLP}&\parbox{10mm}{\centering \bfseries CNN}&\parbox{10mm}{\centering \bfseries UNet}&\parbox{10mm}{\centering \bfseries GT}\\
\hline
Params & 2.02M & 2.93M & 2.03M& 2.52M\\
\hline
$R^{2}$ & 0.995& 0.985 &0.982 &0.871 \\
\hline
MAE\textsubscript{avg,100G} & 0.0099 & 0.0182 & 0.0197&0.0583\\
\hline
MAE\textsubscript{avg,2SRF} & 0.0080 & 0.0149 & 0.0180&0.0521\\
\hline
\end{tabular}
\end{table}

\begin{table}[t]
\caption{Evaluation on the 8-shaped-inductor-based XFMR Dataset}
\centering
\label{evaluation_8shape}
\begin{tabular}{|c|c|c|c|c|}
\hline
\parbox{20mm}{\centering \bfseries {Trained Model}}&\parbox{10mm}{\centering \bfseries MLP}&\parbox{10mm}{\centering \bfseries CNN}&\parbox{10mm}{\centering \bfseries UNet}&\parbox{10mm}{\centering \bfseries GT}\\
\hline
Params & 2.02M & 2.93M & 2.03M& 2.49M\\
\hline
$R^{2}$ & 0.992& 0.986 &0.976 &0.925 \\
\hline
MAE\textsubscript{avg,100G}& 0.0108 & 0.0147 & 0.0191&0.0312\\
\hline
MAE\textsubscript{avg,2SRF}& 0.0060 & 0.0081 & 0.0119&0.0217\\
\hline
\end{tabular}
\end{table}


\subsection{Evaluation of Frequency-domain Self-transfer Learning}

\begin{figure}[t]
\centering
\includegraphics[trim=0 0 0 0, clip, width=3.5in]{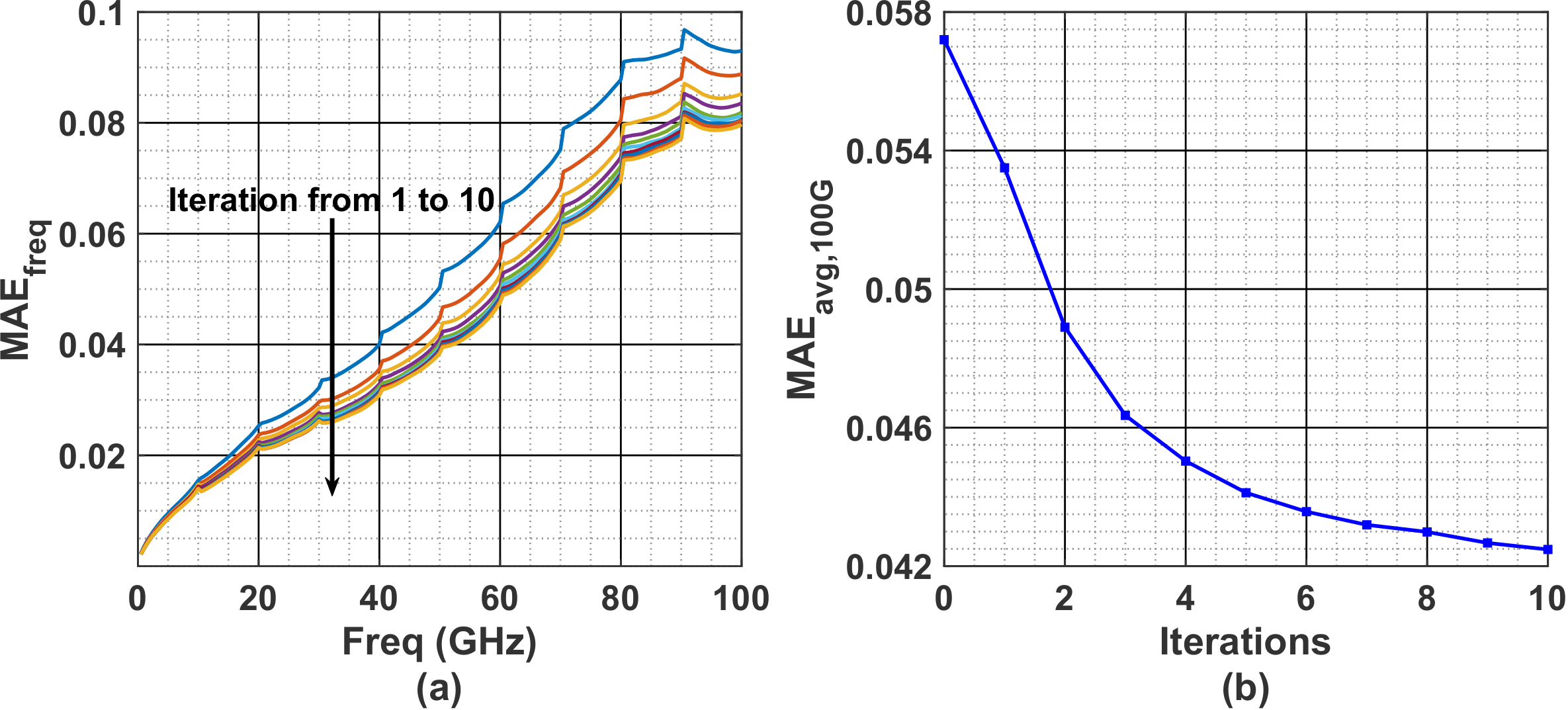}
\vspace{-2em}
\caption{(a) MAE\textsubscript{freq} in each iteration, and (b) MAE\textsubscript{avg,100G} vs. iterations.}
\label{fig:ftr_iteration}
\end{figure}

To evaluate the accuracy improvement enabled by our proposed self-transfer learning technique, we conducted an experiment on the generic $M$:$N$ XFMR dataset. Fig. \ref{fig:ftr_iteration}(a) plots how MAE\textsubscript{freq} evolves over 10 iterations, and Fig. \ref{fig:ftr_iteration}(b) summarizes the corresponding MAE\textsubscript{avg,100G} when the number of frequency sub-bands ($N\textsubscript{band}$) was set to 10.
These results clearly demonstrate the effectiveness of iterative self-transfer learning in improving model accuracy.

\begin{figure}[!t]
\centering
\includegraphics[trim=0 0 0 0, clip, width=3.5in]
{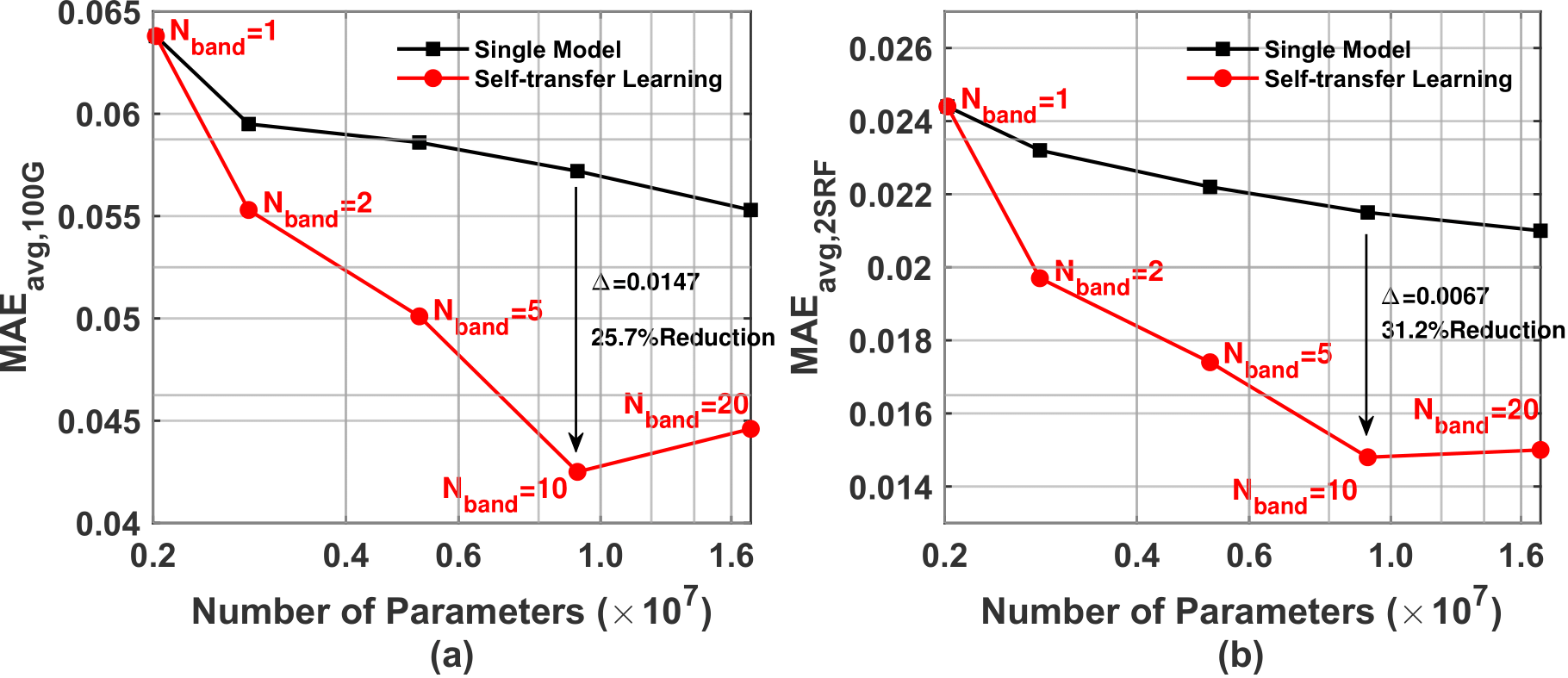}
\vspace{-1.5em}
\caption{Evaluation of self-transfer learning against $N\textsubscript{band}$.}
\vspace{-1.5em}
\label{fig:ftr_bands}
\end{figure}

Fig. \ref{fig:ftr_bands} further explores the impact of varying $N\textsubscript{band}$. When $N\textsubscript{band}$ is small, each sub-model is separated by a large frequency difference, thus limiting the effectiveness of inter-band knowledge transfer. On the other hand, when $N\textsubscript{band}$ becomes too large, each sub-model spans a narrow frequency range, reducing the frequency diversity needed to train each sub-model reliably. These results suggest that there exists an optimal $N\textsubscript{band}$, which is 10 in this experiment. At this setting, MAE\textsubscript{avg,2SRF} drops from 0.0215 to 0.0148, a 31\% reduction in error compared to the baseline MLP-based model with the same number of model parameters. 

Table \ref{evaluation_ftr} presents a comparison of model accuracy with and without the proposed frequency-domain self-transfer learning across all datasets, using the best-performing baseline model for each dataset. With the self-transfer learning technique, a consistent model accuracy improvement of more than 30\% in MAE\textsubscript{avg,2SRF} is achieved. These improved models are then applied to the inverse design tasks presented in the following Section.

\begin{table}[t]
\caption{Model Accuracy Comparison with and without the Proposed Frequency-domain Self-transfer Learning}
\centering
\label{evaluation_ftr}
\begin{tabular}{|c|c|c|c|c|}
\hline
\parbox{20mm}{\centering \bfseries {Dataset}}&\parbox{12mm}{\centering \bfseries $1:1$}&\parbox{12mm}{\centering \bfseries $M:N$}&\parbox{12mm}{\centering \bfseries Parallel}&\parbox{12mm}{\centering \bfseries 8-Shaped}\\
\hline
\parbox{20mm}{\centering \bfseries {Trained Model}}&\parbox{12mm}{\centering \bfseries GT}&\parbox{12mm}{\centering \bfseries UNet}&\parbox{12mm}{\centering \bfseries MLP}&\parbox{12mm}{\centering \bfseries MLP}\\
\hline
MAE\textsubscript{avg,2SRF}, w/o& 0.0099 & 0.0149 &0.0061 &0.0043\\
\hline
MAE\textsubscript{avg,2SRF}, w/&0.0065  & 0.0103 &0.0029 &0.0021\\
\hline
Error Reduction&\textbf{34.3\%} & \textbf{30.9\%}& \textbf{52.4\%}&\textbf{51.2\%}\\
\hline
\end{tabular}
\vspace{-2em}

\end{table}

\section{Inverse Design}
\label{sec:page style}

We illustrate the inverse design of XFMRs by assigning an impedance-matching task as our design target, as impedance matching is a foundational requirement in almost all RFIC designs \cite{HuLNA,yaolongIMS}. The following sub-sections present our cost function, model accuracy requirement, and several exemplar demonstrations. 

\subsection{Impedance-matching Task and Its Cost Function}
\begin{figure}[!t]
\centering
\vspace{0.1em}
\includegraphics[trim=1 1 1 1, clip, width=3.5in]{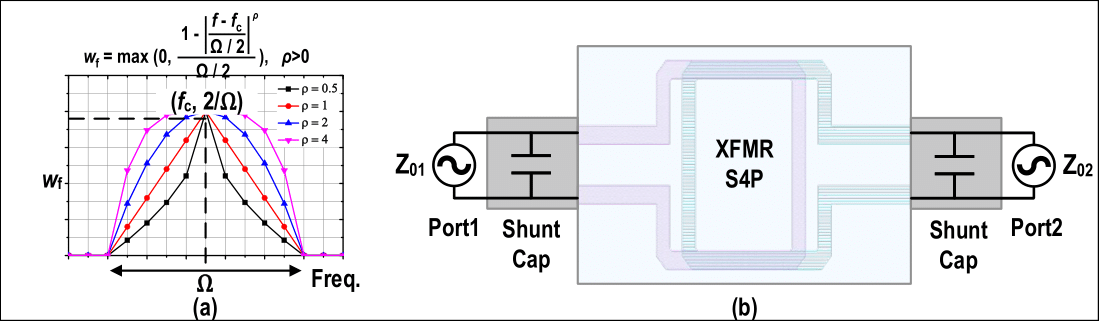}
\vspace{-2em}
\caption{(a) Window function $w_f$. (b) Adding shunt capacitors for matching.}
\label{fig:wf_def}
\end{figure}

In this work, we focus on the differential-mode operation of XFMRs, as RFICs usually employ differential circuits \cite{Outphasing} to enable robustness against common-mode noise (e.g., noise on supply and ground lines). The design objective is to determine the optimal XFMR layout based on the target source impedance $Z_{01}$, load impedance $Z_{02}$, center frequency $f_c$, and bandwidth $\Omega$.
To evaluate inverse design quality, we define a custom cost function, as
\vspace{-1.5em}
\begin{equation}\label{cost_function}
    \resizebox{0.8\hsize}{!}{$
J_S=w_0A+\sum_{f\in\Omega}w_f\Bigg(w_1|\Gamma_{in}|+w_2\bigg(1-|L|\bigg)\Bigg)
$}
\end{equation}
Here, $w_0,~w_1,~w_2$ are weights assigned to the layout area (A), input reflection coefficient ($\Gamma_{in}$), and loss ($L$), respectively. $\Gamma_{in}$ and $L$ can be readily derived from the S-parameters \cite{steer_2019}.  In practice, a well-designed impedance-matching circuit should present $|\Gamma_{in}|<$ --10 dB, and $L$ should be as close to 0 dB as possible \cite{pozar}. $w_f$ is a frequency-domain window function, defined and illustrated in Fig. \ref{fig:wf_def}(a). By adjusting the window index $\rho$, the bandwidth profile of the cost function can be shaped to accommodate different bandwidth specifications. Additionally, we incorporate shunt capacitors into the XFMR-based matching network [Fig. \ref{fig:wf_def}(b)] to compensate reactive components and achieve broadband impedance transformation, a common practice in RFIC designs \cite{XiaohanJSSC2024}.

\vspace{-1em}

\begin{figure}[!t]

\centering
\includegraphics[width=3.6in]{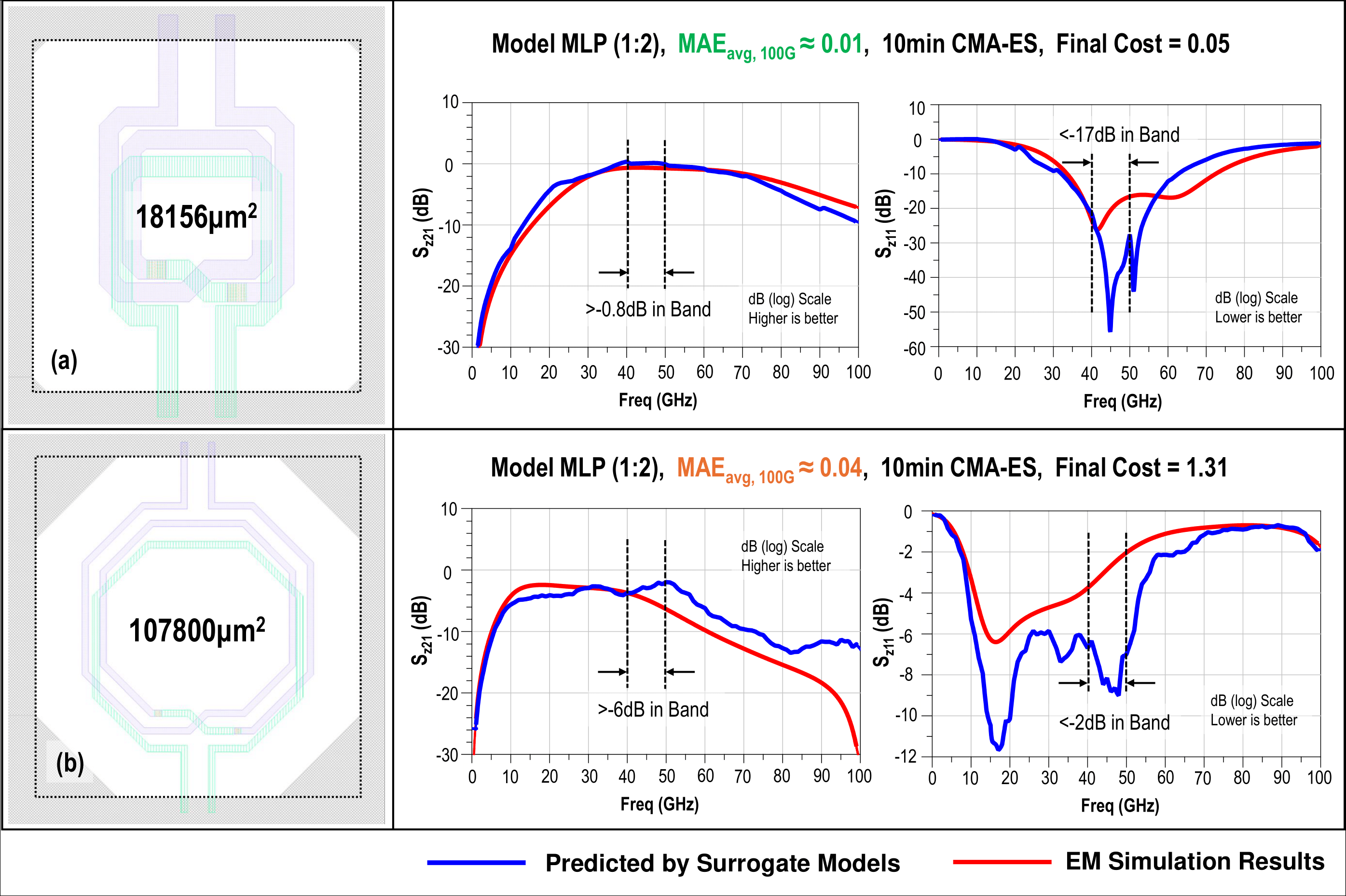}
\vspace{-1.50em}
\caption{Comparison of inverse design quality for a typical impedance-matching task using the 1:2 XFMR topology with different levels of accuracy. Design target: $f_c=45~\text{GHz},\Omega=10~\text{GHz},\rho=1$, $Z_{01}=(40-50j)~\Omega$, $Z_{02}=(150+80j)~\Omega$.}
\vspace{-1.50em}
\label{fig:mae_comp}
\end{figure}

\begin{figure*}[!t]
  \centering
  \includegraphics[width=0.9\textwidth]{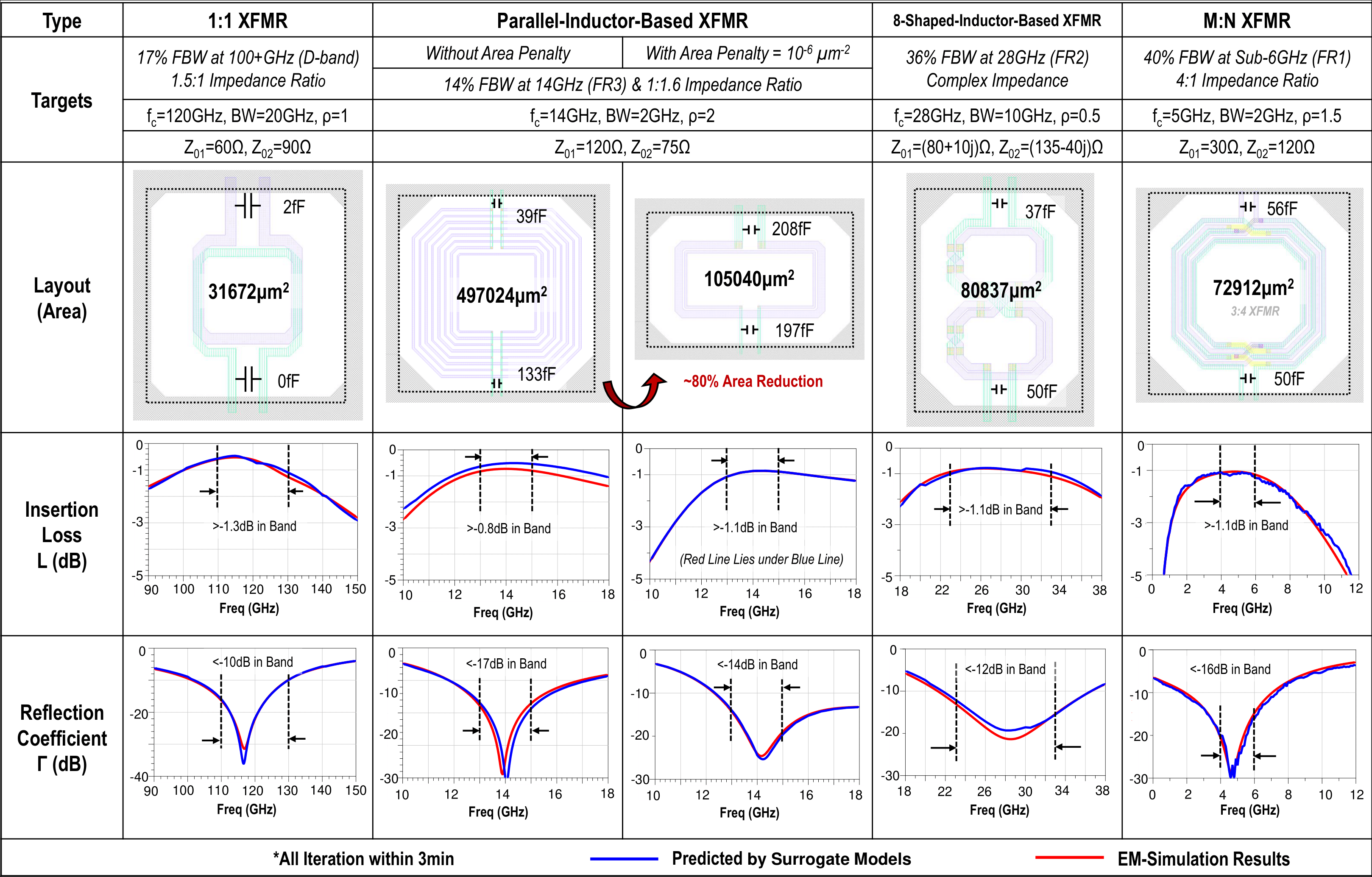}
  \caption{Four inverse design examples using XFMRs for impedance matching, each with a different XFMR topology and a different application scenario.}
  \label{fig:inv_design}
\end{figure*}

\subsection{Inverse Design Reliability vs. Surrogate Model Accuracy}

To evaluate the reliability of inverse design using ML surrogate models, we first conducted experiments with models of different levels of accuracy. Fig. \ref{fig:mae_comp} shows a representative example using the $1$:$2$ XFMR topology. Using a highly accurate surrogate model with an MAE\textsubscript{avg,100G} of about 0.01 [Fig. \ref{fig:mae_comp}(a)], the optimization target is successfully achieved. On the other hand, with a less accurate model with a high MAE\textsubscript{avg,100G} of about 0.04 [Fig. \ref{fig:mae_comp}(b)], the optimization task fails, and there exists an obvious discrepancy between EM-simulated and ML-model-predicted performance. Based on this result and similar experiments we conducted, \emph{an MAE\textsubscript{freq} $<$0.01 at the targeted frequency is generally required} to produce meaningful and trustworthy inverse designs. This also justifies our proposed self-transfer learning technique, which greatly improves the model accuracy to meet this threshold.




\subsection{Inverse Design Examples with Different XFMR Topologies}

In this sub-section, we present four inverse design examples (Fig. \ref{fig:inv_design}), each showcasing a different XFMR topology and targeting different application scenarios with different specifications in terms of center frequency, bandwidth, and impedance transformation ratio. The selected center frequencies of these examples cover a variety of use cases, spanning 5G communication bands (in FR1 and FR2) and beyond-5G communication bands (in FR3 and D-band). 

We adopt the Covariance Matrix Adaptation Evolution Strategy (CMA-ES) algorithm \cite{cmaes} in this work, which dynamically adjusts both the covariance matrix and step size during optimization. 
In repeated experiments, using 4 CPU cores, our framework consistently converged to near-optimal cost values within three minutes.

\section{Conclusion}

This paper presents a comprehensive evaluation of ML surrogate modeling and its application to the inverse design of XFMRs in RFICs. Specifically, we investigated four ML surrogate models, including MLP, CNN, UNet, and GT, on the same dataset and across multiple XFMR topologies. 
Additionally, we developed a frequency-domain self-transfer learning technique that significantly enhances model accuracy.

Our experiments showed that achieving an MAE\textsubscript{freq} below 0.01 is essential to enable trustworthy inverse design. We also validated our end-to-end inverse design process using multiple impedance-matching tasks and further explored the area-performance trade-off. With CMA-ES optimization, all the experimental results demonstrated fast convergence and high-quality inverse designs from specs to GDS.

The surrogate modeling and inverse design approaches presented in this paper not only highlight the potential of AI-assisted design automation for RFICs but also offer a practical guide for RFIC designers who are seeking to integrate ML into their workflows.

\newpage



\bibliographystyle{IEEEtran}
\bibliography{IEEEexample}
\end{document}